\documentclass[sigconf]{acmart}

\usepackage{amsmath}
\usepackage{amsfonts}
\usepackage{multirow}
\usepackage{booktabs}
\usepackage{tabularx}
\usepackage{enumitem}
\usepackage{subfloat}
\usepackage{epstopdf}
\usepackage[labelfont=bf,textfont=bf]{caption}

\usepackage{amsmath}
\usepackage{booktabs}
\usepackage{tabularx}
\usepackage{adjustbox}
\usepackage{mdwlist}
\usepackage{amsfonts}
\usepackage{multirow}
\usepackage{array}
\usepackage{epstopdf}
\usepackage{enumitem}
\usepackage{subfigure}
\usepackage{bm}
\usepackage{dcolumn}
\usepackage{array}
\usepackage{enumitem}
\usepackage{listings}
\usepackage{xcolor}
\usepackage{wrapfig}

\usepackage{algorithm}  
\usepackage{algpseudocode}
\usepackage{epsfig}
\usepackage{bbm}
\usepackage{makecell}
\usepackage{pifont}
\usepackage{bbding}
\usepackage{color}
\usepackage{svg}

\newcommand{\stitle}[1]{\vspace{0.8mm} \noindent {\bf #1}}

\definecolor{mycustompurple}{RGB}{154, 36, 79} % 定义自己的颜色
\usepackage[utf8]{inputenc}
\usepackage{hyperref} % 引入超链接宏包
\hypersetup{
    colorlinks=true,            % 激活链接颜色，去掉链接边框
    linkcolor=red,              % 文档内部链接颜色（如图表等引用）
    citecolor=black,            % 文献引用链接颜色
    filecolor=mycustompurple,   % 文件链接颜色
    urlcolor=magenta            % 外部URL链接颜色
}

\AtBeginDocument{%
  }

\copyrightyear{2025}
\acmYear{2025}
\setcopyright{acmlicensed}
\acmConference[SIGIR '25] {Proceedings of the 48th International ACM SIGIR Conference on Research and Development in Information Retrieval}{ July 13--18, 2025}{Padua, Italy.}
\acmBooktitle{Proceedings of the 48th International ACM SIGIR Conference on Research and Development in Information Retrieval (SIGIR '25), July 13--18, 2025, Padua, Italy}
\acmISBN{979-8-4007-1592-1/25/07}
\acmDOI{10.1145/3726302.3730015}

\begin{document}

\title{IVCR-200K: A Large-Scale Multi-turn Dialogue Benchmark for Interactive Video Corpus Retrieval}

    \author{Ning Han}
    \authornote{Equal contribution.}
    \affiliation{
     \institution{Xiangtan University}
     \country{Hunan, China}
    }
    \email{hanninginf@gmail.com}
    
    \author{Yawen Zeng}
    \authornotemark[1]
    \affiliation{
      \institution{Hunan University}
      \country{Hunan, China}
    }
    \email{yawenzeng11@gmail.com}
    
    \author{Shaohua Long}
    \affiliation{
     \institution{Xiangtan University}
     \country{Hunan, China}
    }
    \email{longshaohua@smail.xtu.edu.cn}
    
    \author{Chengqing Li}
    \authornote{Corresponding author.}
    \affiliation{
     \institution{Xiangtan University}
     \country{Hunan, China}
    }
    \email{DrChengqingLi@gmail.com}
    
    \author{Sijie Yang}
    \affiliation{
     \institution{Hunan University}
     \country{Hunan, China}
    }
    \email{ysijie42@gmail.com}
    
    \author{Dun Tan}
    \affiliation{
     \institution{Hunan University}
     \country{Hunan, China}
    }
    \email{tandun1049@gmail.com}
    
    \author{Jianfeng Dong}
    \affiliation{
     \institution{Zhejiang Gongshang University}
     \country{Zhejiang, China}
    }
    \email{dongjf24@gmail.com}
    
    \author{Jingjing Chen}
    \affiliation{
     \institution{Fudan University}
     \country{Shanghai, China}
    }
    \email{chenjingjing@fudan.edu.cn}

\renewcommand{\shortauthors}{Ning Han et al.}

\begin{abstract}
In recent years, significant developments have been made in both video retrieval and video moment retrieval tasks, which respectively retrieve complete videos or moments for a given text query. These advancements have greatly improved user satisfaction during the search process. However, previous work has failed to establish meaningful \textbf{``interaction''} between the retrieval system and the user, and its one-way retrieval paradigm can no longer fully meet the personalization and dynamic needs of at least 80.8\% of users. 
In this paper, we introduce the Interactive Video Corpus Retrieval (IVCR) task, a more realistic setting that enables multi-turn, conversational, and realistic interactions between the user and the retrieval system. To facilitate research on this challenging task, we introduce IVCR-200K, a high-quality, bilingual, multi-turn, conversational, and abstract semantic dataset that supports video retrieval and even moment retrieval. Furthermore, we propose a comprehensive framework based on multi-modal large language models (MLLMs) to help users interact in several modes with more explainable solutions. The extensive experiments demonstrate the effectiveness of our dataset and framework. The datasets, codes, and leaderboards are available at: \url{https://ivcr200k.github.io/IVCR}.
\end{abstract}




%% information and the body of the document, and typically spans the page.

\maketitle

\section{Introduction}\label{sec:intro}

\begin{figure*}[t]
    \center
    \includegraphics[width=0.8\textwidth]{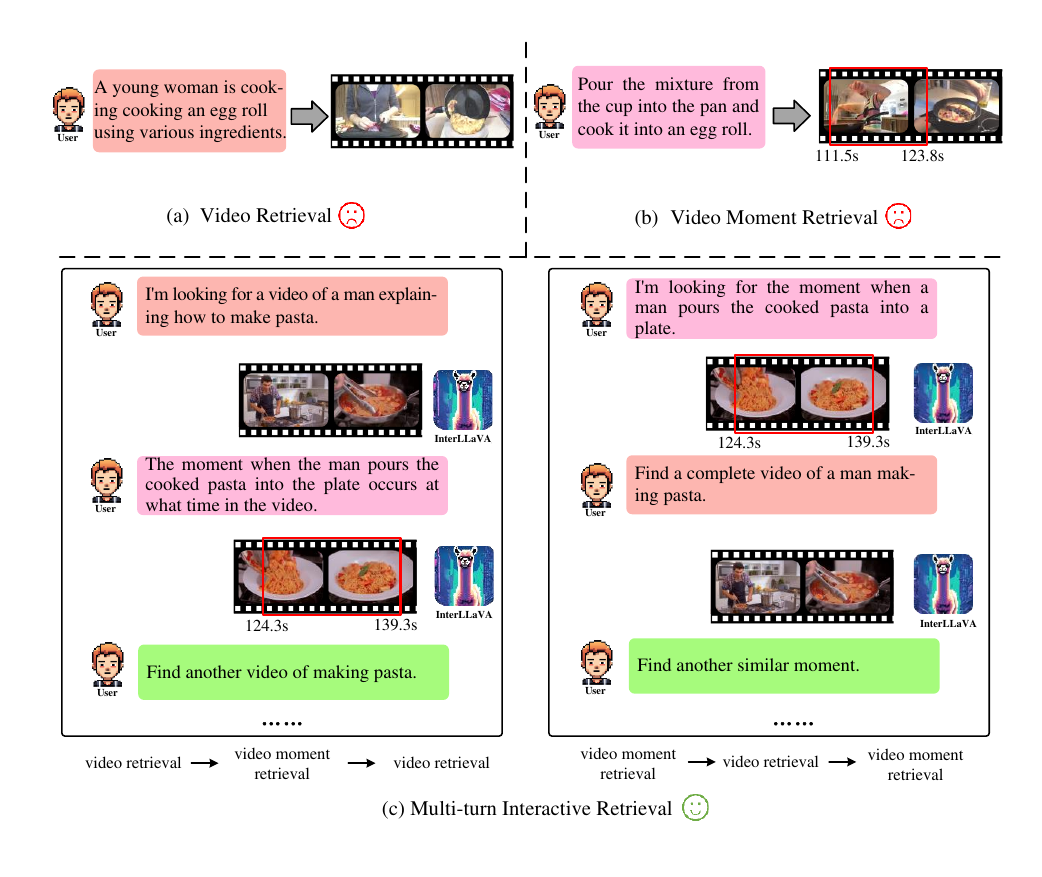}
    \vspace{-0.3cm}
    \caption{Visualization of the video retrieval, moment retrieval, and our multi-turn interactive retrieval.}
    \label{fig:example}
    \vspace{-0.2cm}
\end{figure*}

With the rapid proliferation of video platforms such as YouTube and TikTok, an ever-increasing number of videos are being produced daily, underscoring the significance of the video retrieval task in the multimodal field \citep{dong2021dual,han2021fine,zeng2021multi}. Typically, users employ descriptive sentences, and the retrieval system \citep{xu2016msr,luo2021clip4clip} sorts by matching textual descriptions and visual videos, ultimately returning the user's preferred videos, as depicted in Figure~\ref{fig:example}(a). At a more granular level, as shown in Figure \ref{fig:example}(b), researchers have proposed the video moment retrieval task \citep{gao2017tall,zeng2022prompt}, which utilizes textual descriptions to retrieve a small moment within the complete video. These tasks significantly enhance user satisfaction during the search process.

\begin{figure*}[t]
    \centering
    %\vspace{\baselineskip}
    \subfigure[]{
    \begin{minipage}[t]{0.45\linewidth}
    \centering
    \includegraphics[width=\linewidth]{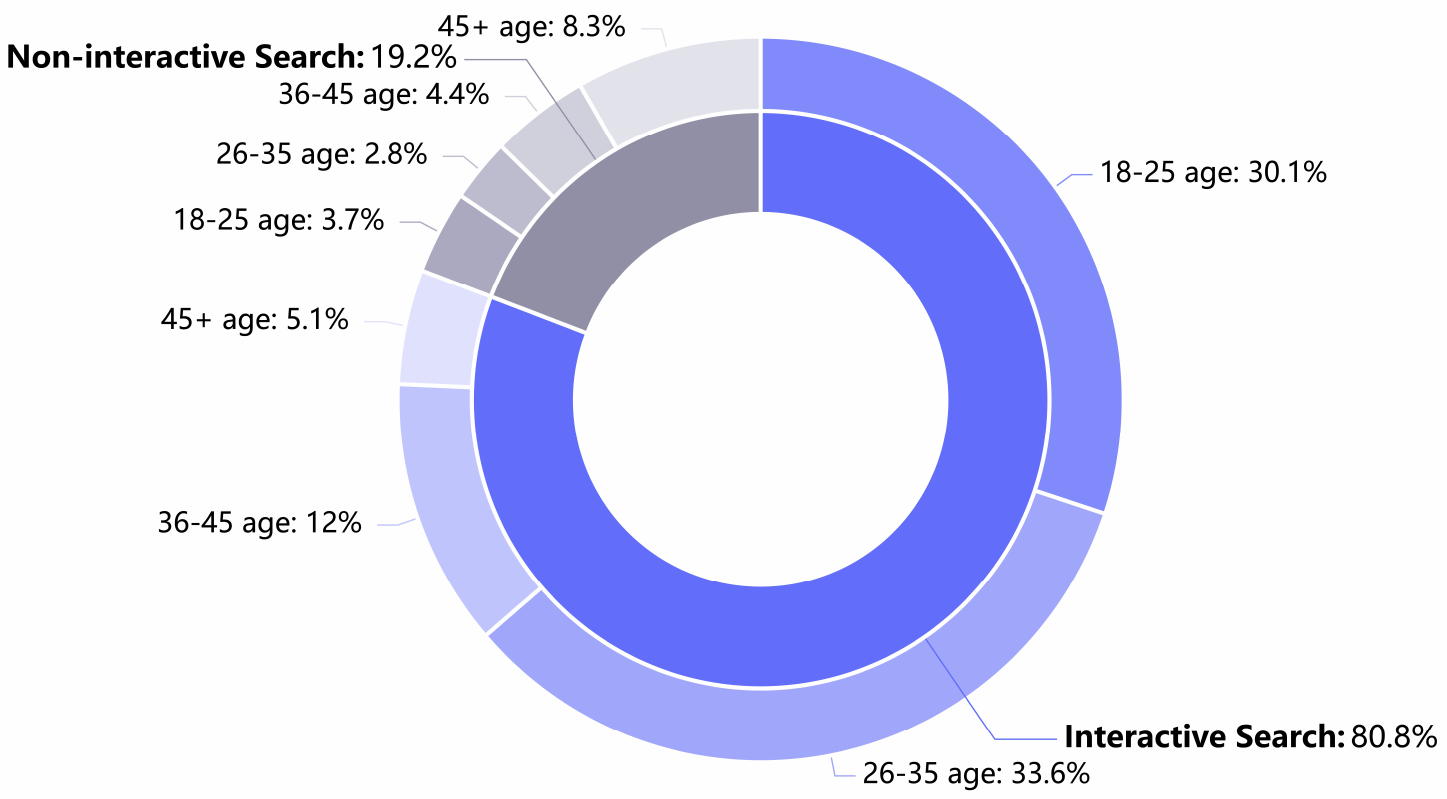}
    %\caption{Questionnaire results regarding interactive search.}
    \end{minipage}%
    }%
    \subfigure[]{
    \begin{minipage}[t]{0.35\linewidth}
    \centering
    \includegraphics[width=\linewidth]{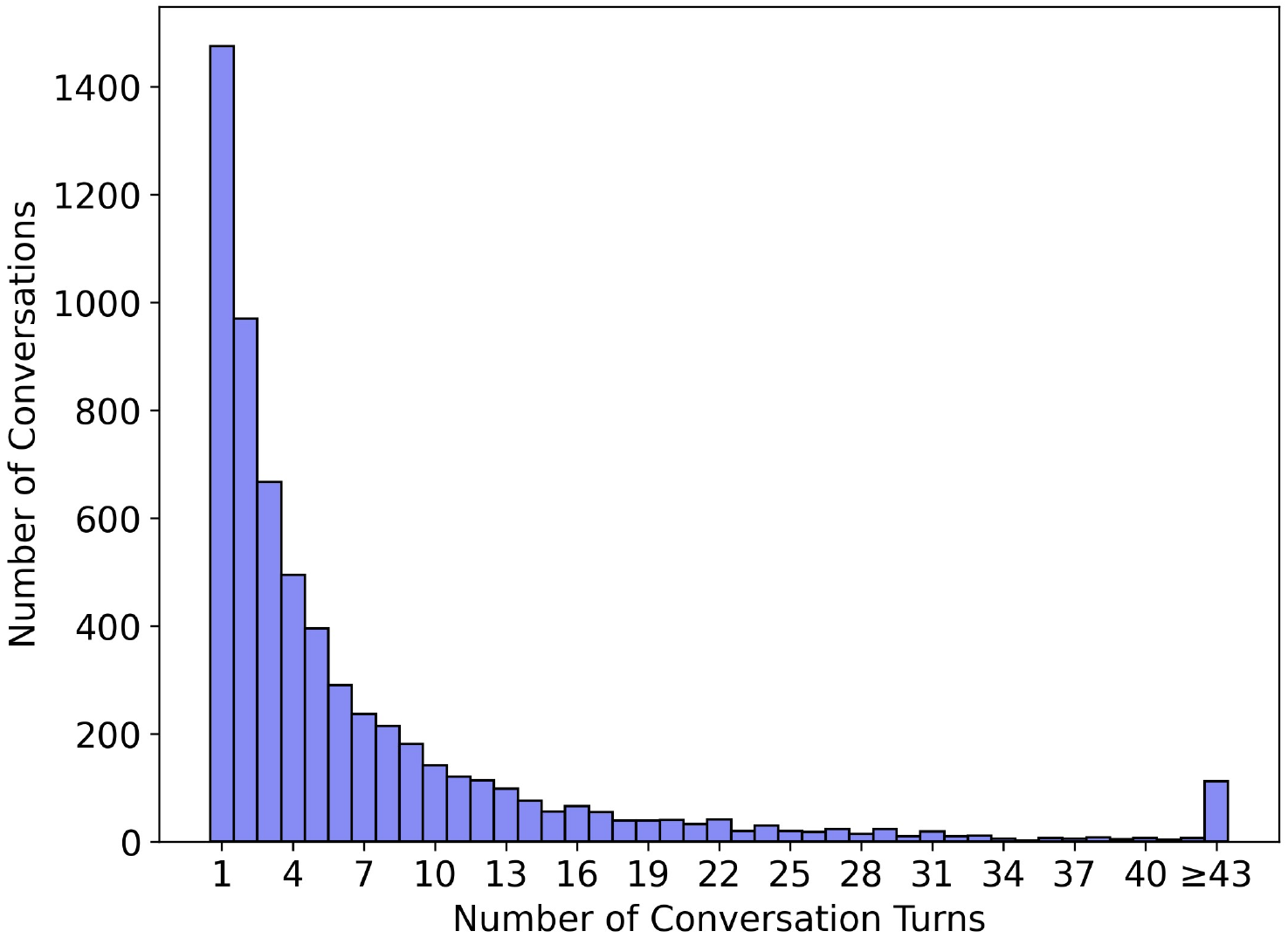}
    %\caption{The distribution of interaction turns in ShareGPT.}
    \end{minipage}%
    }%
    \centering
    \vspace{-0.3cm}
    \caption{Investigation of User Search Behavior, Feedback, and Interaction Turns in ShareGPT. Users demonstrate a pronounced inclination towards interactive search and harbor high expectations regarding interaction rounds.}
    \label{fig:example2}
    \vspace{-0.2cm}
\end{figure*}

However, the majority of video retrieval systems operate in a ``one-way'' manner, which may not fully cater to users' personalized and dynamic preferences. This ``one-way'' approach inhibits user interaction with the system, resulting in every request from the user needing to be rewritten. It is a common phenomenon that users desire \textbf{``multi-turn interaction''} with systems. To delve deeper into this phenomenon, we devised a questionnaire regarding user search behavior, depicted in Figure \ref{fig:example2}. A striking $80.8\%$ of respondents expressed a preference for interactive search functionality. Similarly, within the ShareGPT\footnote{https://sharegpt.com/} conversation dataset, the average interaction round between users and the chat system stands remarkably high at $7.27$. Moreover, the questionnaire indicates that interactive demands exhibit intricate behavioral patterns, as illustrated in Figure \ref{fig:example}(c): 
1) Long2Short: Keep looking for clips within the long videos that have already been scanned;
2) Short2Long: Search full-length videos based on known short videos;
3) Analogous: When the user inputs ``I would like to watch a movie similar to this clip", the system should be able to provide a video with similar content. Therefore, drawing on these observations, we believe that implementing an interactive retrieval system holds significant value \citep{ma2022interactive,maeoki2020interactive}, despite the challenges posed by complex user behaviors. Through multi-turn interaction with users, the system can adapt to individual preferences, furnishing more personalized retrieval outcomes. However, researchers have yet to thoroughly investigate this practical issue, which would better meet users’ needs.

\input
\setlength{\tabcolsep}{4mm}  % Adjusts the horizontal padding between columns
\renewcommand\arraystretch{1.2}  
\begin{table*}
    \centering
    \caption{Comparison of IVCR-200K and other existing video-language datasets.}
    \vspace{-0.2cm}
    \label{tab:comparison}
    \newsavebox{\tabb}
    \begin{lrbox}{\tabb}
    \resizebox{\textwidth}{!}{
    \begin{tabular}{c|c|c|c|c|c|c}
    \toprule
    Dataset   & Multi-turn                    & Dialogue                      & Real interaction                & Videos & Queries & Language      \\ \hline
    
    MSR-VTT\citep{xu2016msr}      & {\color{red}\XSolidBrush}   & {\color{red}\XSolidBrush}   & {\color{red}\XSolidBrush}  & 7,180 & 200K & English       \\ \hline
    
    MSVD\citep{chen2011collecting}           & {\color{red}\XSolidBrush}   & {\color{red}\XSolidBrush}   & {\color{red}\XSolidBrush}  & 1,790 &70K & English \\ \hline
    
    LSMDC\citep{rohrbach2017movie}    & {\color{red}\XSolidBrush}   & {\color{red}\XSolidBrush}   & {\color{red}\XSolidBrush}  & 200 &118K  & English \\    \hline  
    
    ActivityNet Captions\citep{krishna2017dense}           & {\color{red}\XSolidBrush}   & {\color{red}\XSolidBrush}   & {\color{red}\XSolidBrush}  & 20,000 & 100K  & English           \\    \hline        
                                           
     VATEX\citep{wang2019vatex}         & {\color{red}\XSolidBrush}   & {\color{red}\XSolidBrush}   & {\color{red}\XSolidBrush} & 41,250 & 825K    & English, Chinese      \\ \hline
     
     HowTo100M\citep{miech2019howto100m}      & {\color{red}\XSolidBrush}   & {\color{red}\XSolidBrush}   & {\color{red}\XSolidBrush}  & 1.221M   & 136M  & English    \\ \hline
    
    Charades-STA\citep{gao2017tall}    & {\color{red}\XSolidBrush}   & {\color{red}\XSolidBrush}   & {\color{red}\XSolidBrush}    & 6,670  & 16,128  & English      \\ \hline
    
    DiDeMo\citep{anne2017localizing}    & {\color{red}\XSolidBrush}   & {\color{red}\XSolidBrush}   & {\color{red}\XSolidBrush}    & 10,464   & 41K  & English    \\ \hline
    
    TVQA\citep{lei2018tvqa}        & {\color{red}\XSolidBrush}   & {\color{green}\CheckmarkBold}   & {\color{red}\XSolidBrush}  & 21,793 & 152,545  & English           \\    \hline   
    
    AVSD\citep{alamri2019audio}  & {\color{green}\CheckmarkBold}   & {\color{green}\CheckmarkBold}    & {\color{red}\XSolidBrush}    & 11,816   & 118,160  & English     \\ \hline
    
     \textbf{IVCR-200K} & {\color{green}\CheckmarkBold} & {\color{green}\CheckmarkBold} & {\color{green}\CheckmarkBold}  & 12,516 & 201,631 & English, Chinese \\
    \bottomrule
    \end{tabular}}
    \end{lrbox}
    \scalebox{0.9}{\usebox{\tabb}} 
\end{table*}

Formally, we introduce the Interactive Video Corpus Retrieval (IVCR) task. We define the ``interactive'' as meeting the following requirements: \textbf{1) Multi-turn}. This multi-turn interaction will extend the connection between the user and the search system. This process involves several interaction modes, including video retrieval-only, moment retrieval-only, video-first-then-moment, moment-first-then-video, and creating a new topic for retrieval. \textbf{2) Free dialogue}. Users perform queries in natural language \citep{alayrac2022flamingo,dai2024instructblip}, and the retrieval system should explain the returned results in natural language form, which is more explainable and user-friendly. Furthermore, existing multimodal retrieval datasets primarily contain low-level descriptive descriptions (e.g., ``There are three dogs on the green lawn''), which do not align with the high-level abstract semantics used by users in real-world scenes (e.g., ``Kung Fu movie where men and women fight''). \textbf{3) Real interaction}. The pioneers create simulated environments to generate interactive data \citep{ma2022interactive}, but we emphasize that only truly collecting data sent by users can optimize a better search experience.

Unfortunately, at present, there is no available dataset or reliable framework to support this task of interactive video corpus retrieval, as shown in Table \ref{tab:comparison}. \textbf{1) Dataset}. Existing video retrieval datasets are inadequate for multi-turn interaction scenarios, such as ActivityNet Captions \citep{krishna2017dense} and DiDeMo \citep{anne2017localizing}, which are single-turn datasets. Therefore, we propose an innovative interactive retrieval dataset, IVCR-200K, which is a bilingual, multi-turn, conversational, and high-quality dataset with abstract semantic content, designed to support video retrieval and even moment retrieval. \textbf{2) Framework}. Existing retrieval methods are insufficient for these conversational scenarios. For instance, solutions like CLIP4Clip \citep{luo2021clip4clip} and 2D-TAN \citep{zhang2020learning} are discriminative models that cannot perform dialogue generation. Inspired by recent advances in multi-modal large language models \citep{li2023blip,ren2024timechat}, we combine their multi-turn dialogue, semantic understanding, and other capabilities to support users' interaction modes with a more explainable solution, named InterLLaVA. Extensive experiments demonstrate the effectiveness of our dataset and framework.

\begin{figure*}[t]
    \center
    \includegraphics[width=0.85\textwidth]{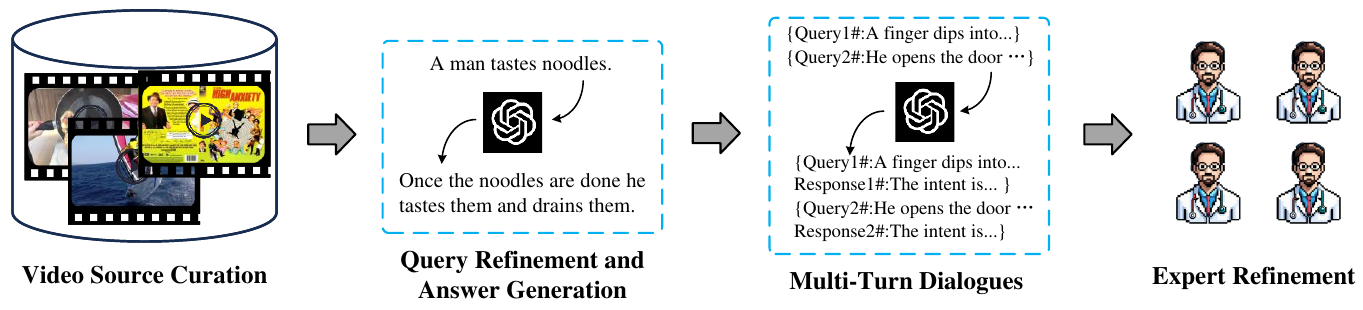}
    \vspace{-0.3cm}
    \caption{The pipeline of the dataset collection.}
    \label{fig:dataset}
    \vspace{-0.3cm}
\end{figure*}

The main contributions are summarized as follows: 
1) To the best of our knowledge, this is the first work to introduce the ``interactive'' video corpus retrieval task, which effectively aligns users' multi-turn behavior in real-world scenarios and significantly enhances user experience; 
2) We introduce a dataset and an accompanying framework. Notably, the IVCR-200K dataset is a high-quality, bilingual, multi-turn, conversational, and abstract semantic dataset designed to support video and moment retrieval. The InterLLaVA framework leverages multi-modal large language models (MLLMs) to enable multi-turn dialogue experiences between users and the retrieval system.

\section{Related Work}\label{sec:related} 

\stitle{Video Retrieval Dataset.} In recent years, with the vigorous development of the digital video new media market and continuous technological innovation, the scale of datasets related to video retrieval has rapidly expanded. For example, Xu et al. \citep{xu2016msr} constructed a video understanding dataset MSR-VTT, which contains 10K clips and 20K different text descriptions corresponding to various categories. MSVD \citep{chen2011collecting} is also a widely used dataset in video retrieval, containing 1,970 videos, with each video having approximately 40 associated sentences. Rohrbach et al. \citep{rohrbach2017movie} developed the LSMDC, comprising 200 movies and 128,118 sentences, which is widely used for cross-model retrieval between video and text. Krishna et al. \citep{krishna2017dense} built a large-scale dataset ActivityNet Captions for dense captioning events, which contains 20k videos and a total of 100k descriptions, each with its unique start and end times. In comparison, Howto100M \citep{miech2019howto100m} comprises over 23k different visual tasks and 136 million video clips from 1.22M instructional web videos with narration, which is the largest video retrieval dataset. Wang et al. \citep{wang2019vatex} constructed a large-scale multilingual video description dataset VATEX, which contains over 41,250 videos along with 825,000 captions in both English and Chinese. Gao et al. \citep{gao2017tall} built a dataset called Charades-STA, which augments the existing Charades \citep{sigurdsson2016hollywood} dataset by adding sentence temporal annotations for temporal activity localization via language. However, these datasets are primarily designed to support video retrieval or video moment retrieval research, rather than interactive video corpus retrieval, and therefore do not meet the personalized and dynamic retrieval needs of users.

In this paper, we build the IVCR-200K dataset with 12K videos and more than 200K sentences covering 36 categories. To the best of our knowledge, IVCR-200K is the first and the most extensive video dataset for interactive video corpus retrieval. The dataset is a crucial step in developing deep learning-based methods. We hope the dataset will inspire further efforts in the task of interactive video corpus retrieval.

\stitle{Video Retrieval.}
Recently, numerous video datasets have been released for various video-language understanding tasks. In Table \ref{tab:comparison}, we present a statistical comparison of the IVCR-200K dataset with ten video datasets used for video retrieval tasks. Video retrieval aims to retrieve relevant videos from a set of video candidates given a text query \citep{han2024efficient,han2023bic}. We categorize related methods into pre-extracted multi-modality fusion approaches that integrate motion, audio, and face cues to boost retrieval performance \cite{liu2019use, gabeur2020multi, wang2021t2vlad}, joint text–video pre-training techniques that learn end-to-end from raw videos paired with text \cite{yan2021video, li2022align, ge2022bridging}, and CLIP-based frameworks that adapt a pre-trained CLIP backbone for text-video retrieval \cite{luo2021clip4clip, gorti2022x, liu2022ts2}. As an extension of video retrieval, the video moment retrieval task aims to identify specific clips or moments within a video based on a given textual query \citep{gao2017tall,he20read}. Pioneering works have explored various technical avenues, including attention-based retrieval \cite{liu2018attentive, wang2021visual}, reinforcement learning \cite{zeng2023rewardtlg, he20read}, visual-language pretraining \cite{zeng2023temporally, ren2024timechat}, among others. These studies have enhanced the retrieval system's service capabilities. However, further development is required to meet the multi-turn interactive needs of users. 

\stitle{Interactive Retrieval.}
The concept of interactive retrieval has long been proposed as a means of combining human-machine learning techniques for multimedia content search \citep{thomee2012interactive,snoek2008videolympics}. Currently, only a few works \citep{madasu2022learning, maeoki2020interactive, ma2022interactive,liang2023simple} have explored this task. For example, Madasu et al. \citep{madasu2022learning} and Maeoki et al. \citep{maeoki2020interactive}adopt a dialogue-based approach, utilizing a series of video-related questions and answers generated by different models as retrieval queries. Furthermore, Ma et al. \citep{ma2022interactive}develop a user simulation for intelligent multimedia applications to enable precise video segment search through human-computer interaction. The technical challenges in modeling multi-turn dialogue retrieval have contributed to the slow development in this direction.

\stitle{Large language Models.} With the breakthroughs in generative artificial intelligence, the way humans interact with machines has changed \citep{min2023recent,IBSurvey24,zheng2024judging}. Researchers have extended large language models to the visual perception domain, developing a series of large language models with multimodal information processing capabilities, such as Flamingo \citep{alayrac2022flamingo}, BLIP-2 \citep{li2023blip}, and LLaVA \citep{liu2024visual} for image processing, and Sora, Video LLaMA \citep{zhang2023video}, and Video Chat \citep{li2023videochat} for video understanding. Specifically, for interactive cross-modal video retrieval, future interactive video retrieval systems should function as ``search assistants'', engaging in genuine and coherent multi-round dialogues with users.

\section{Interactive Video Corpus Retrieval Dataset}\label{sec:dataset}

\subsection{Dataset Collection and Annotation}

To implement an interactive video retrieval system, we constructed a multi-turn, conversational dataset comprising 226,102 interactions sourced from 5 video repositories. This dataset encompasses functionalities such as video retrieval, video moment retrieval, and natural dialogue.

\begin{figure*}[t]
    \centering
    %\vspace{\baselineskip}
    %\subfigure[]{
    \begin{minipage}[t]{0.3\linewidth}
    \centering
    \includegraphics[width=\linewidth]{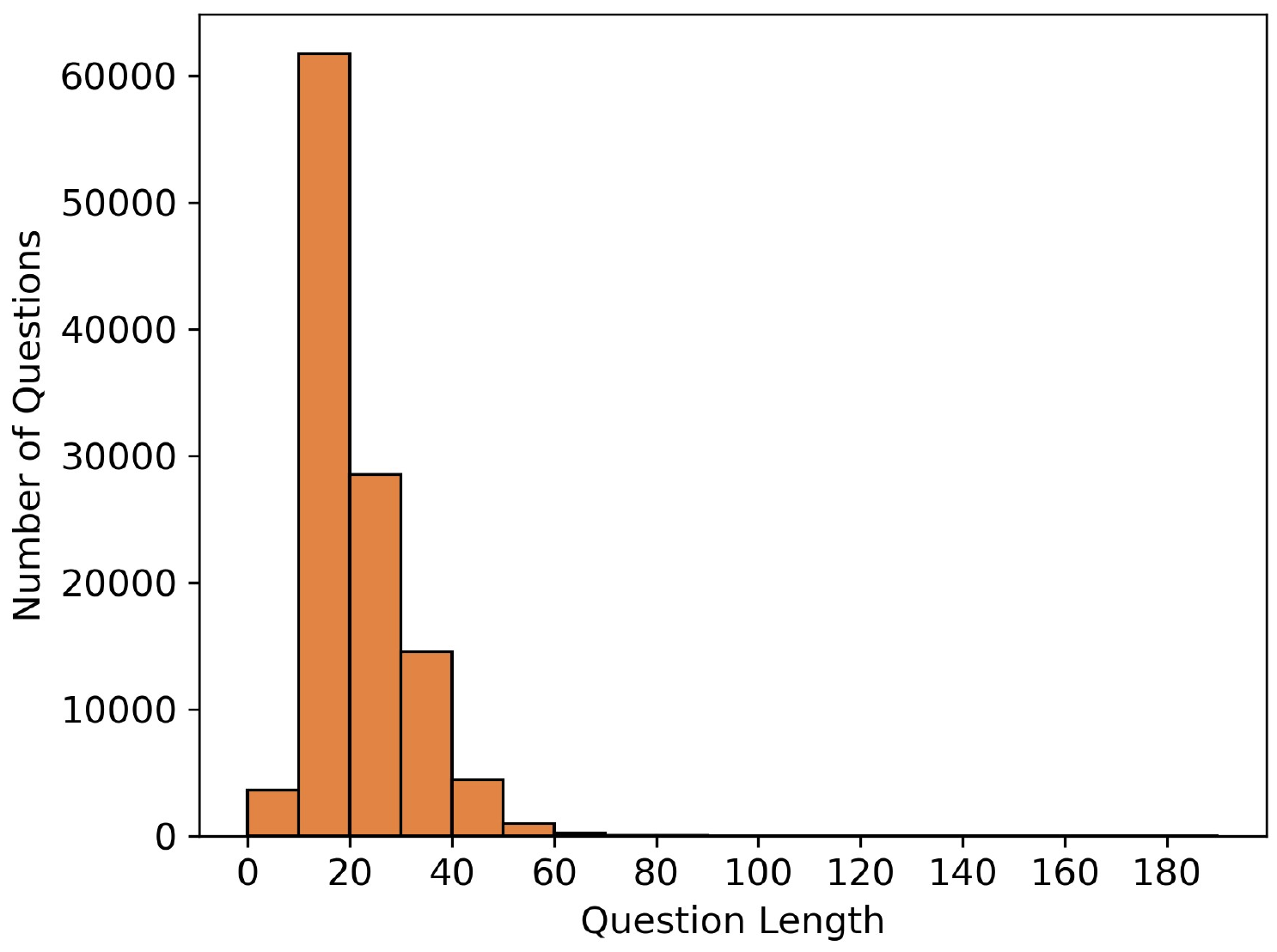}
    %\caption{fig1}
    \end{minipage}%
    %}%
    %\subfigure[]{
    \begin{minipage}[t]{0.3\linewidth}
    \centering
    \includegraphics[width=\linewidth]{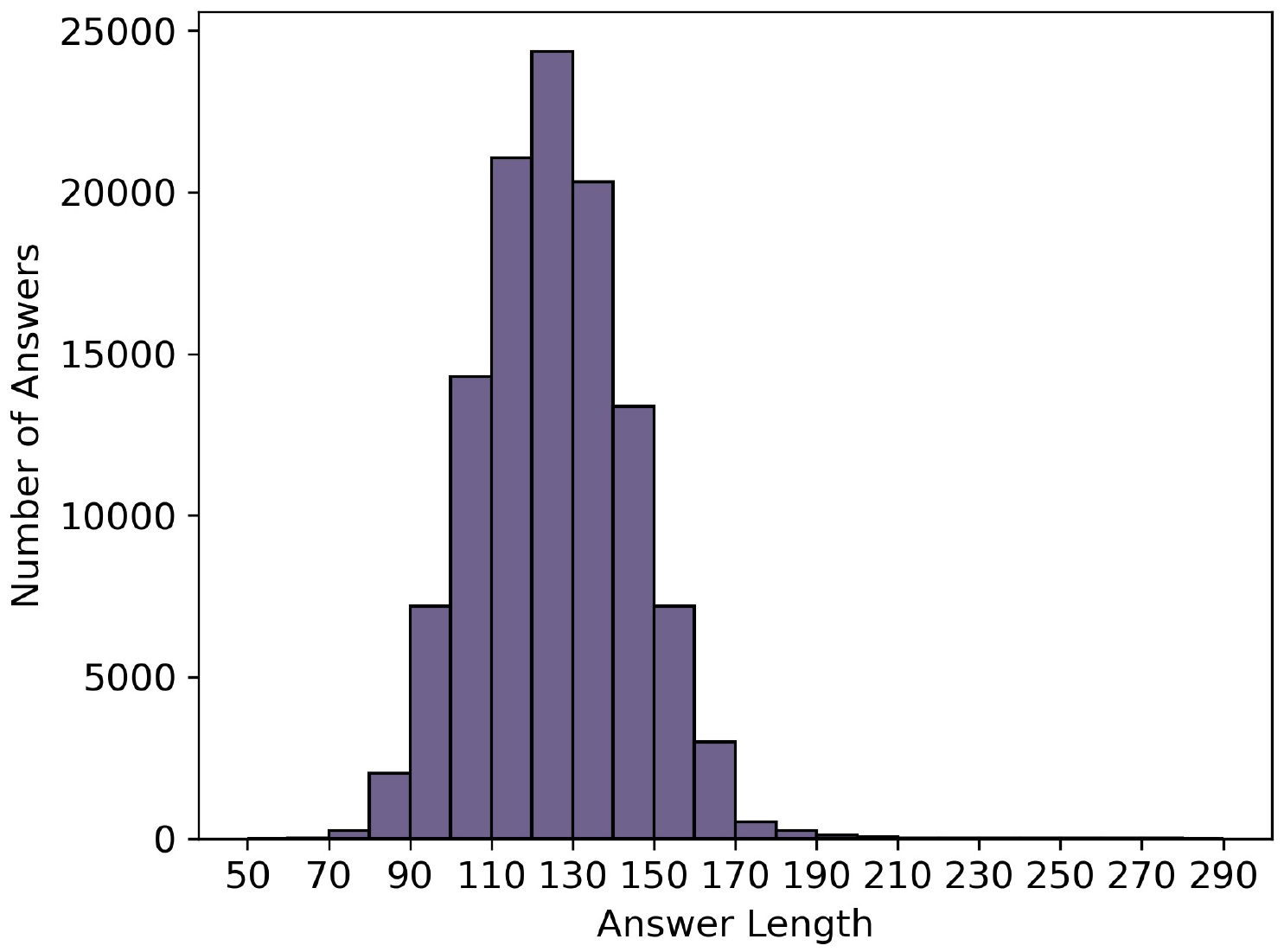}
    %\caption{fig2}
    \end{minipage}%
    %}%
    %\subfigure[]{
    \begin{minipage}[t]{0.303\linewidth}
    \centering
    \includegraphics[width=\linewidth]{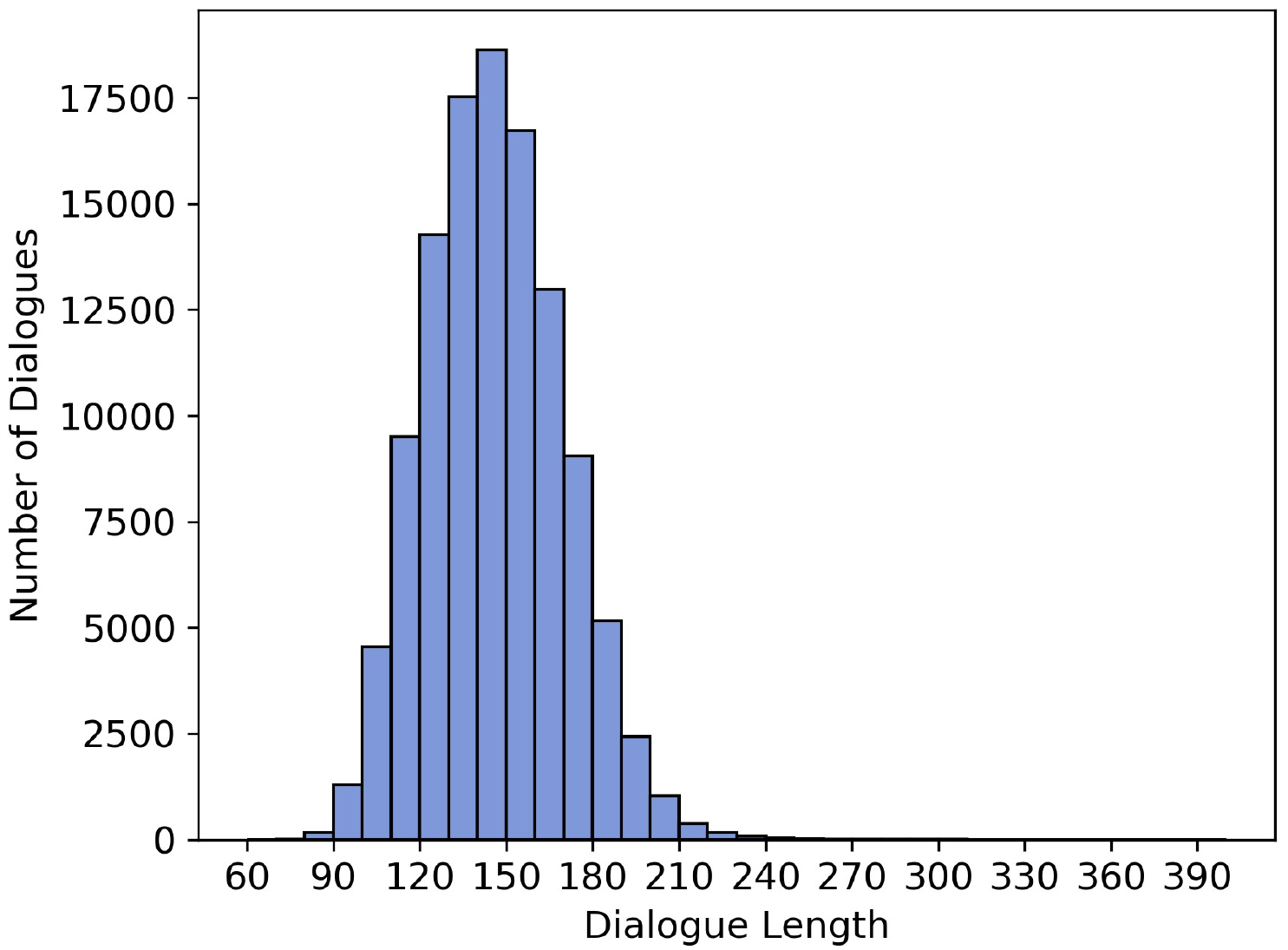}
    %\caption{fig3}
    \end{minipage}%
    %}%
    \centering
    \caption{Distribution of question lengths, answer lengths, and dialogue lengths.}
    \label{fig:dialogue}
    %\vspace{-0.2cm}
\end{figure*}

\begin{figure*}[t]
    \centering
    %\vspace{\baselineskip}
    %\subfigure[]{
    \begin{minipage}[t]{0.298\linewidth}
    \centering
    \includegraphics[width=\linewidth]{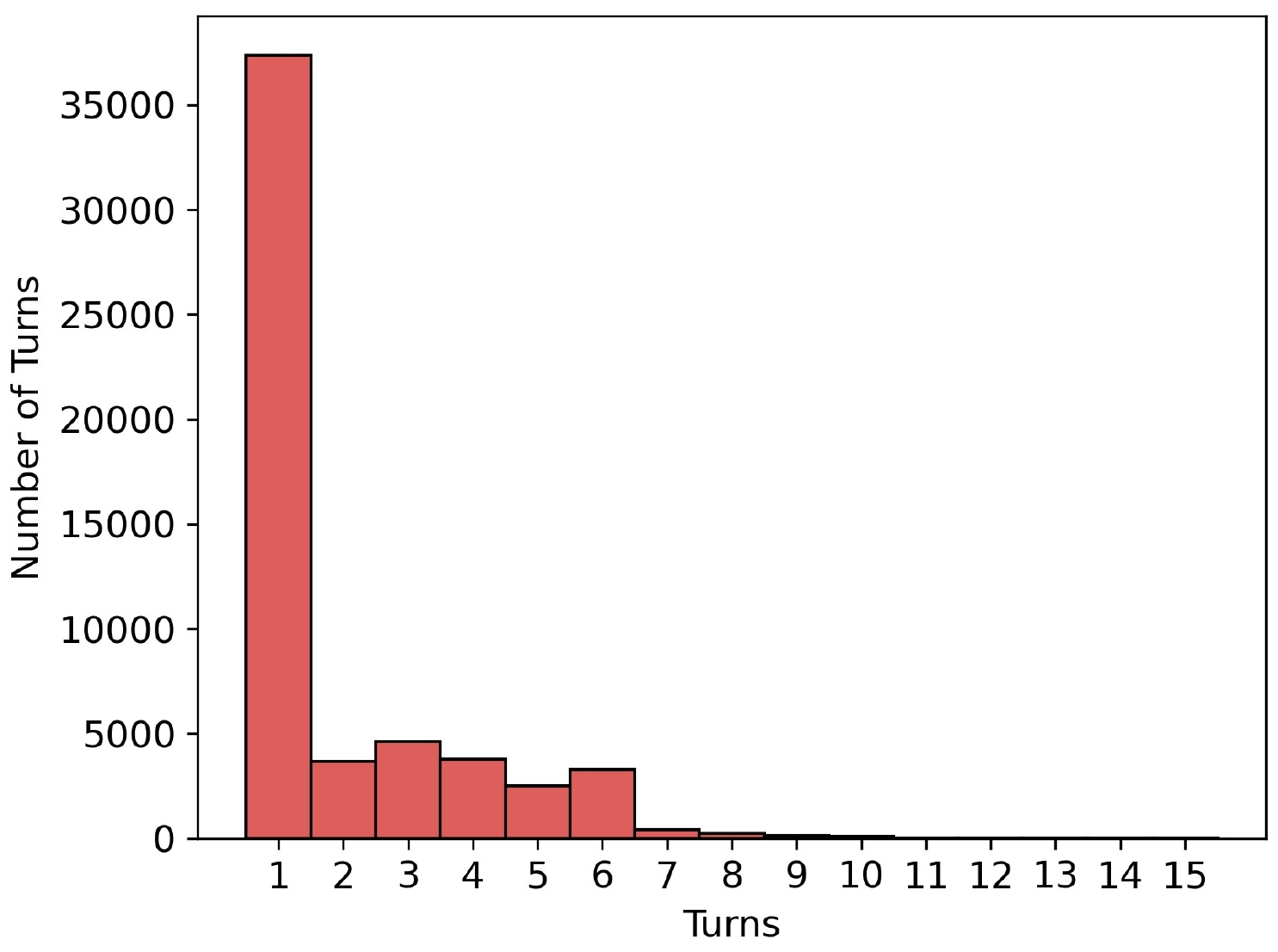}
    %\caption{fig1}
    \end{minipage}%
    %}%
    %\subfigure[]{
    \begin{minipage}[t]{0.3\linewidth}
    \centering
    \includegraphics[width=\linewidth]{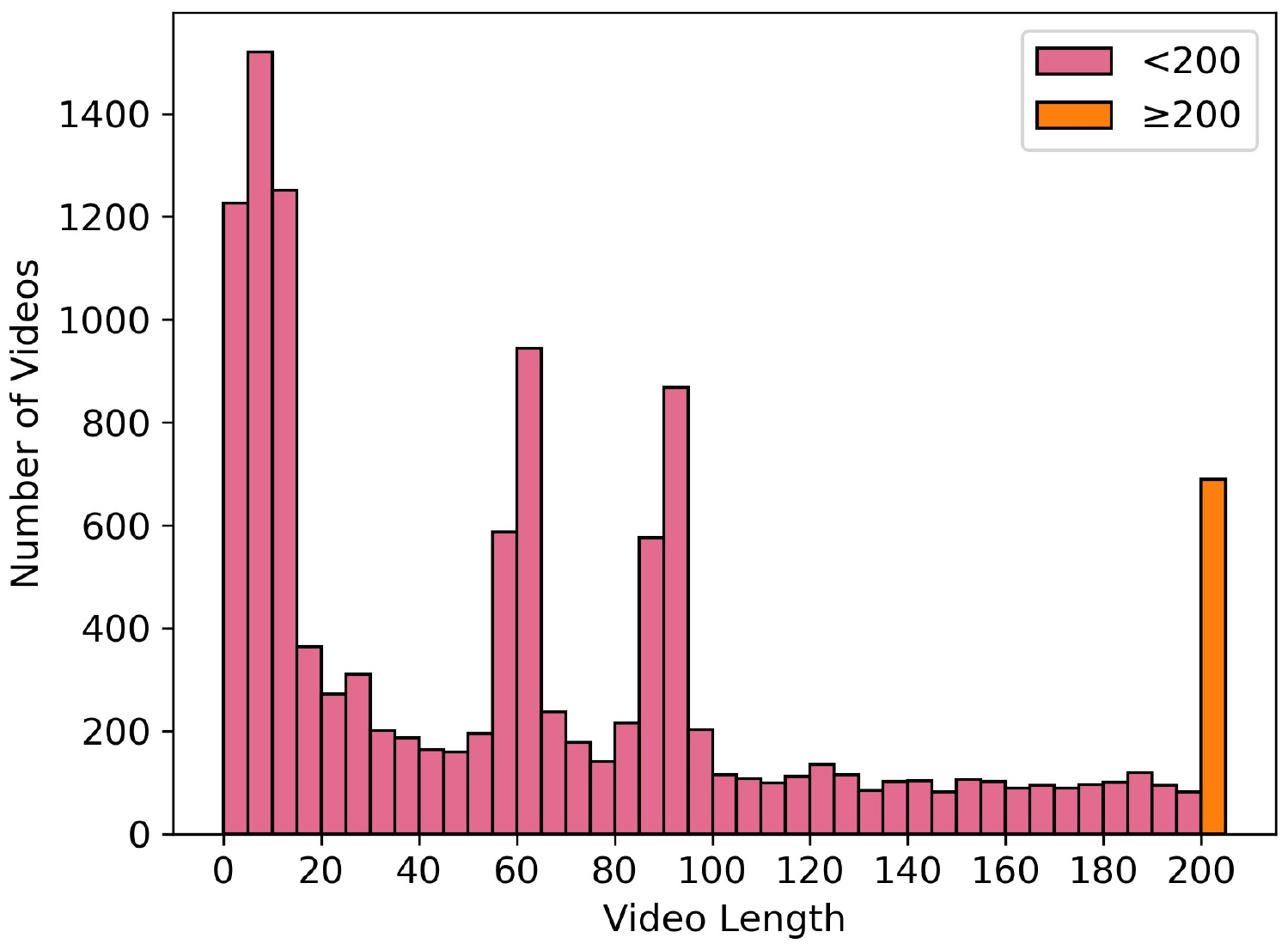}
    %\caption{fig2}
    \end{minipage}%
    %}%
    %\subfigure[]{
    \begin{minipage}[t]{0.3\linewidth}
    \centering
    \includegraphics[width=\linewidth]{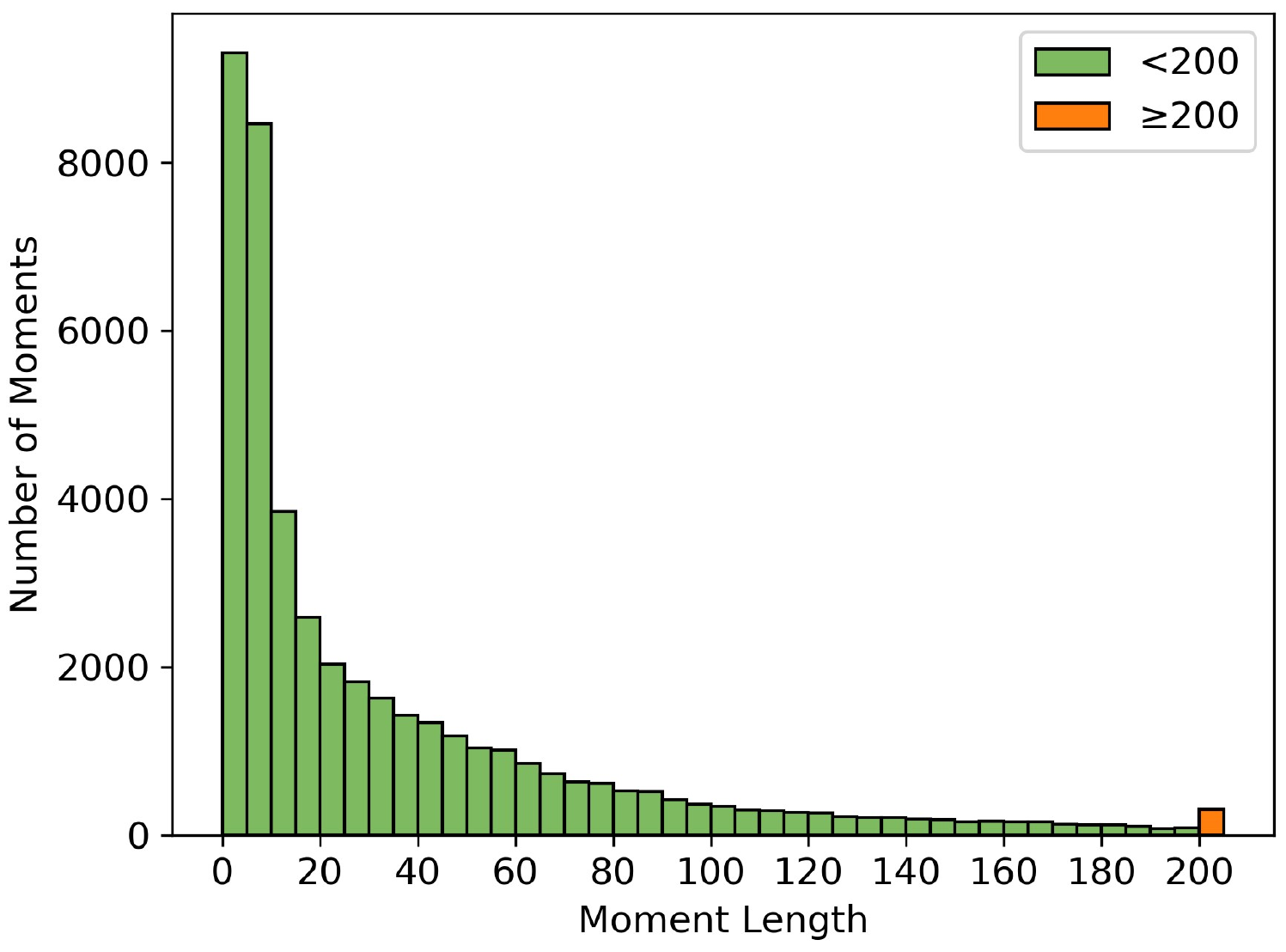}
    %\caption{fig2}
    \end{minipage}%
    %}%
    \centering
    \caption{Distribution of turn lengths, video lengths, and moment lengths.}
    \label{fig:videos}
    %\vspace{-0.5cm}
\end{figure*}

Illustrated in Figure \ref{fig:dataset}, we devised a comprehensive collection pipeline:
\begin{itemize}[leftmargin=*]
    \item 1) Video source curation: Initially, we selected video datasets spanning diverse domains such as daily activities, movies, and kitchens, including selections like TVQA \citep{lei2018tvqa}, LSMDC \citep{rohrbach2017movie}, Activi-tyNet Captions \citep{krishna2017dense}, DiDeMo \citep{anne2017localizing}, MSR-VTT \citep{xu2016msr}, to ensure video source diversity. Subsequently, we filtered out select videos from these five original datasets. Videos featuring isolated actions or events, severe occlusion, or excessively accelerated playback were excluded. Ultimately, 12,516 videos were chosen for inclusion. 
    
    \item 2) Query refinement: Despite the presence of captions or descriptions with the filtered source videos, they often inadequately align with user queries in real-world scenarios. Hence, we employed GPT-4 for query refinement on captions. Specifically, we first combine the captions and user queries as input, and then use GPT-4 to generate user queries that more accurately and closely reflect the substantive content of the video.
    
    \item 3) Multi-turn dialogues: We established various dialogue dynamics, encompassing Long2Short, Short2Long, Long2Long, Short2Short, and Natural Dialogue scenarios. ``Long2Short'' denotes a user's inclination to pinpoint video clips further in the current round, while ``Natural Dialogue'' reflects users' perception of our system as a standard chat robot.
    Notably, while most dialogues consist of concatenated single-round exchanges, we also gathered a limited number of multi-turn dialogues from actual users.
    
    \item 4) Interpretability: To bolster the interpretability of interactive retrieval systems, we utilized GPT-4 to craft responses, encompassing intent understanding, retrieval or localization results, and reasons.
    
    \item 5) Bilingual capability: To broaden the reach of the dataset, we employed a translation model to render the dataset into Chinese.
\end{itemize}

Notably, every output produced by GPT-4 will undergo meticulous scrutiny and refinement by human experts to guarantee the precision of knowledge. Additionally, we implemented a validation process conducted by a review team, focusing on the quality and consistency of annotations provided by different annotators. After all annotations (201,631 sentence-level queries) were completed, the reviewers further examined the annotated data. Ultimately, we acquired a multi-turn, conversational dataset comprising 200K volumes, named IVCR-200K. The entire annotation and review process took approximately five months to complete. 

\begin{figure}[t]
    \center
    \includegraphics[width=0.47\textwidth]{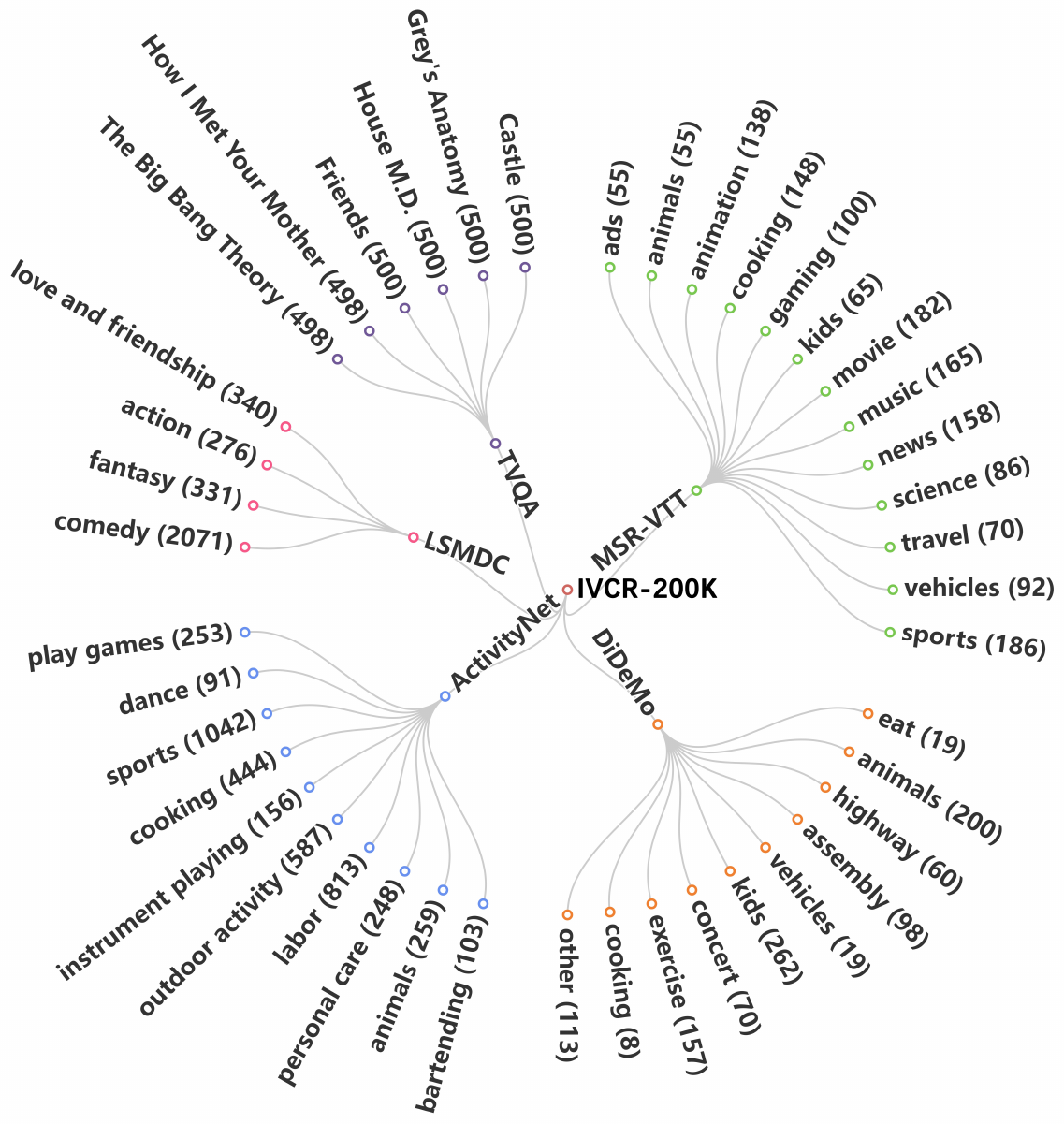}
    \caption{The hierarchical structure of the sources and categories of IVCR-200K.}
    \label{fig:category}
    \vspace{-0.3cm}
\end{figure}

\subsection{Dataset Analysis} 

\stitle{Property Quality.} 
The statistical analysis of property quality for video and textual queries in the IVCR-200K dataset is presented in Figures \ref{fig:dialogue} and \ref{fig:videos}. In Figure \ref{fig:dialogue}, we present the length distribution of questions, answers, and dialogues within IVCR-200K. The average length of questions and answers in IVCR-200K is 20.73 words and 125.03 words, respectively. In contrast, the average length of questions in AVSD \citep{alamri2019audio} is 7.9 words, and the average answer length is 9.4 words. This indicates that the dialogues in the dataset are more verbose and conversational.
Additionally, Figure \ref{fig:videos} shows the distribution of the number of turns in multi-turn dialogues. The total number of dialogue turns is 226,102, which aligns with typical user retrieval behavior. Figure \ref{fig:videos} also presents the length distribution of videos and video moments. The average length of videos is 68.43 seconds, and the average length of video moments is 33.8 seconds, with most video moments being under 60 seconds.

\stitle{Diversity Quality.} We analyzed video sources of IVCR-200K, including different types of videos, to ensure comprehensive diversity (See Figure \ref{fig:category}).

\begin{figure}[t]
    \center
    \includegraphics[width=0.45\textwidth]{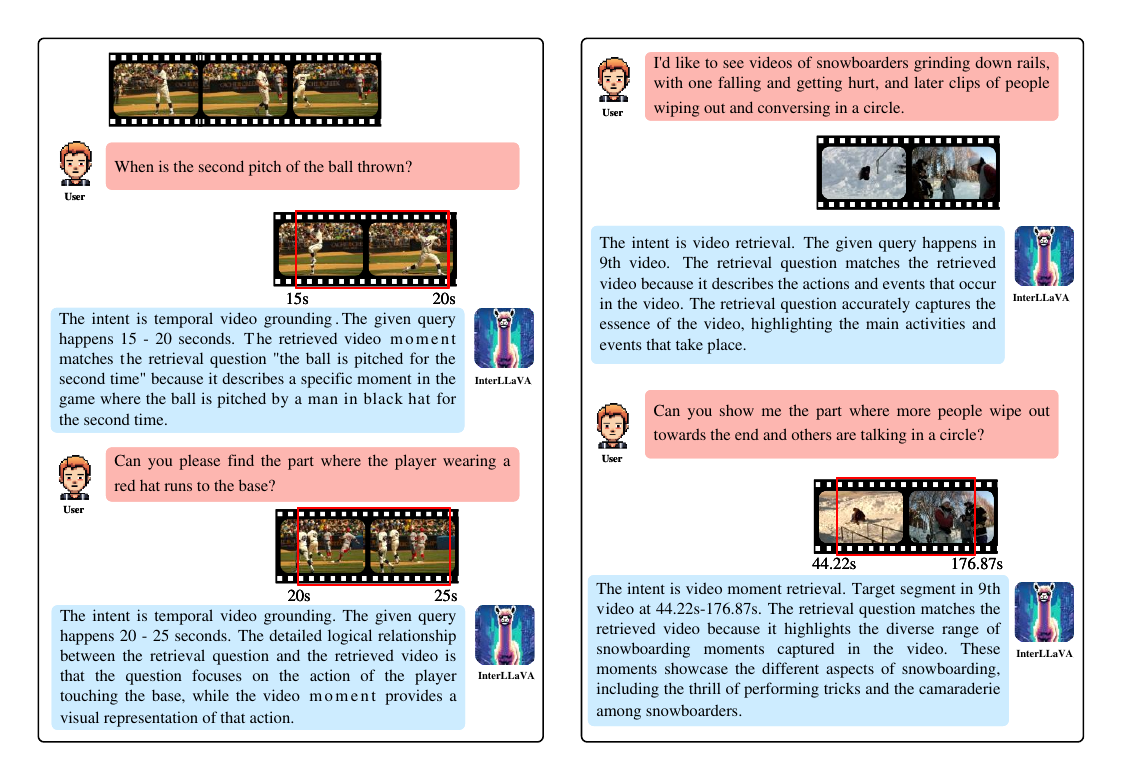}
    \caption{Examples from the IVCR-200K dataset.}
    \label{fig:dialog}
    \vspace{-0.3cm}
\end{figure}

\stitle{Visualization Quality.} We also check some cases as shown in the Figure \ref{fig:dialog}. {More examples are available at this \href{https://ivcr200k.github.io/IVCR}{link}}.

\section{Interactive Video Corpus Retrieval Framework}\label{sec:method}

%Figure\ref{fig:framework} shows the overview of our framework (i.e., InterLLaVA) pipeline. It aims to adapt the pretrained large language model LLaMA-2(7B)\cite{touvron2023llama} to tackle IVCR tasks. Our framework imitates a real user through offering relevance feedback and assessing the retrieval results. In the following, we first introduce the definition of the IVCR task in Section\ref{sec:def}. Then, we delve into the specifics of the three-task processing method in Section\ref{sec:task}.

\subsection{Task Definition}\label{sec:def} 

Let $u_{(\cdot)}$ denotes a user whose historical interactive sequence is $Q = \{q_1, q_2, q_3, q_4, ...\}$, where $q_{(\cdot)}$ represents different textual queries. Formally, the goal of this interactive video corpus retrieval task is to retrieve semantically matched videos or moments in each round $i$, based on the historical interactive sequence $Q_{<i}$. Among them, video moment retrieval requires not only the prediction of the most suitable video $v_j$, but also the prediction of the optimal moment within $v_j$, which includes the start $s$ and end $e$ timestamps. In addition, the interactive video corpus retrieval task is not limited to video retrieval, but also specifically considers the identification and processing of natural dialogue intent.

\subsection{Task Processing}\label{sec:task} 

\begin{figure*}[t]
    \center
    \includegraphics[width=0.83\textwidth]{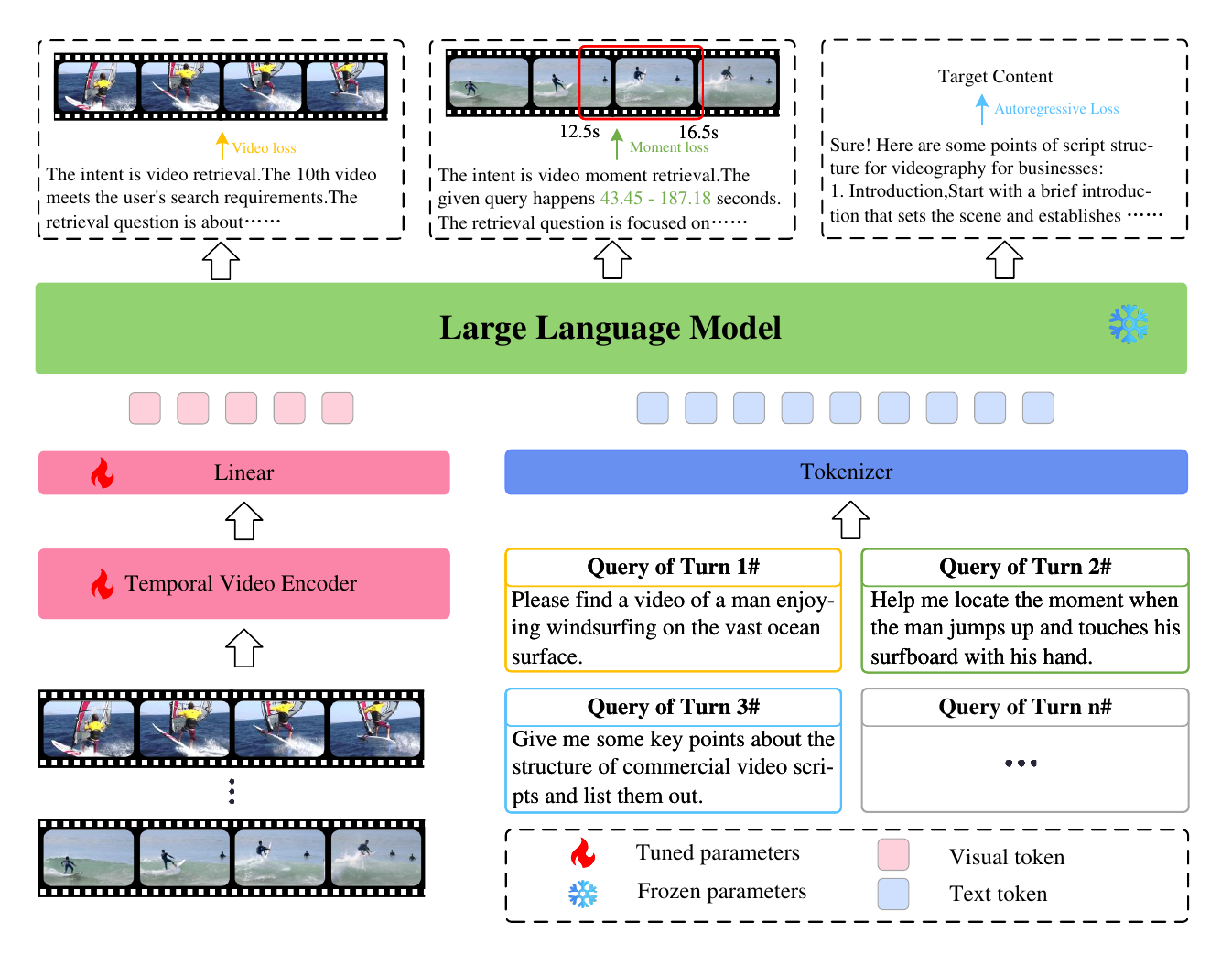}
    \vspace{-0.1cm}
    \caption{An overview of the proposed framework for interactive retrieval.}
    \label{fig:framework}
    %\vspace{-0.1cm}
\end{figure*}

As illustrated in Figure \ref{fig:framework}, InterLLaVA adapts the pretrained large language model LLaMA-2 (7B) \citep{touvron2023llama} to tackle video retrieval, video moment retrieval, and natural dialogue in a multi-turn setting. It takes video and text queries as inputs and outputs video, video moment, and natural dialogue related to textual query intent, while providing interpretable feedback. Specifically, we fine-tuned Inter-LLaVA using instruction-tuning data, which generally consists of video-instruction pairs. Here is an illustrative example, with the underlined part serving a pseudo-instruction:
\begin{center}
\fcolorbox{black}{gray!10}{\parbox{.85\linewidth}{
Video Retrieval:\\
Question: \#\#\#\, Human: <VID>Video ID</VID> <VIDEO> [VIDEOTOKEN] </VIDEO> ... [User Qusetion] \\
Answer: \#\#\#\, Assistant: The intent is video retrieval. The given query happens in [Video ID] video. [Explainable Feedback] \\

Video Moment Retrieval:\\
Question: \#\#\#\, Human: <VID>Video ID</VID> <VIDEO> [VIDEOTOKEN] </VIDEO> [Timestamps] [User Qusetion] \\
Answer: \#\#\#\, Assistant: The intent is video moment retrieval. The given query happens in video [Video ID] at [Start Time] - [End Time] seconds. [Explainable Feedback]
}}
\end{center}

During the instruction fine-tuning of InterLLaVA, text query is first performed using a pre-trained large language model (LLaMA-2 (7B)), which is then concatenated with video and answer prompts to serve as the input for InterLLaVA. The answer prompts include retrieval intent, video, moment cues, and interpretable feedback. Later, the answer prompts are utilized as the “ground truth” of InterLLaVA’s generation. In the following, we elaborate on the implementations of the three tasks. 

\stitle{Video Retrieval.} 
For this task, we propose combining a fast two-tower video model with a multi-modal large language model through a re-ranking mechanism. Specifically, in the first phase, the video retrieval model predicts the top-10 video sequence $V_j$ based on videos and text queries. In the second phase, the top-10 video sequences and text queries are input into a multimodal large language model for re-ranking, outputting the most relevant video $v_j$. This approach retrieves the most pertinent videos efficiently, reduces the memory and computational burden on the language model, and excludes irrelevant content. Note that the first phase involves offline video sequence extraction, while the second phase is trained end-to-end alongside the other tasks.

\stitle{Video Moment Retrieval.} 
For this task, we employ a traditional two-stage retrieval method, utilizing a fast two-tower model for video retrieval and a multi-modal large language model for precise moment localization. Specifically, we implement a two-phase approach. In the first phase, the video retrieval model directly outputs the top-1 video $v_j$. In the second phase, the textual query and the top-1 video are input into a multi-modal large language model to generate a reasonable and coherent response and video moment. To enhance the feature fusion in the time dimension, we adopt a sliding video Q-Former and initialize it from the Video-LLaMA \citep{zhang2023video} checkpoint. Moreover, we perform instruction tuning on the IVCR-200K dataset, which contains timestamp-related and natural dialogue data, to further strengthen InterLLaVA's timestamp localization and natural dialogue capabilities.

\stitle{Training and Inference.}
In training, we implement a two-phase approach. In the first phase, we train a video retrieval model based on the video and text features encoded by CLIP \citep{radford2021learning}, utilizing X-Pool \citep{gorti2022x} as the base model. The video retrieval model acts as a plug-in for the multi-modal large language model, retrieving the top-10 video sequences or the top-1 video. In the second phase, we fine-tune the InterLLaVA with instruction data to achieve instruction following. To better tailor LLaMA for video tasks, we leverage the LoRA \citep{hu2021lora} technique for efficient parameter fine-tuning. To adapt to the IVCR task, we designed a new loss function for training InterLLaVA. For training the large model, we employ a language model loss to generate the target answer $R_a$ with a length of $L_t$. This loss is based on the probability of predicting each word in the answer sequence given the context, such as video tokens $F_v$ and the query tokens $F_q$. It is formulated as
\begin{equation}\label{eqn:loss1}
\begin{aligned}
\mathcal{L}_{M} &= -\log P_{\Theta}(R_a | F_v, F_q)  \\
                &= -\sum_{i=1}^{L_t} \log P_{\Theta}(r_i | R_{a,<i}, F_v, F_q),
\end{aligned}
\end{equation}
where $\Theta$ represents the trainable parameters, and $R_{a,<i}$ refers to the answer tokens preceding the current prediction token $r_i$.

Since our goal is to enhance the large language model's ability for video re-ranking, a direct approach is to optimize the predicted video index directly against the ground-truth video index. Let $v_p$ be the predicted video index, and $v_g$ denote the ground truth video index. The cross-entropy loss function is computed as
\begin{equation}\label{eqn:loss2}
\mathcal{L}_{V} = -\sum_{i=1}^{N} v_{g,i} \log(v_{p,i}),
\end{equation}
where $N$ is the total number of video indices, $v_{g, i}$ is the ground truth probability distribution (with 1 for the correct index and 0 for others), and $v_{p, i}$ is the predicted probability for the $i$-th video index.
Similarly, let $c_p$ be the predicted video moment interval, and $c_g$ denotes the ground truth video moment interval. We force the model to align each predicted moment candidate with the ground truth moment. Our model is trained by an Intersection over Union (IoU) loss \citep{yu2016unitbox} as 
\begin{equation}\label{eqn:loss3}
\mathcal{L}_{C} = 1 - \text{IoU}(c_p, c_g).
\end{equation}
The overall loss function for training the InterLLaVA is the sum of the three losses:
\begin{equation}\label{eqn:loss}
\mathcal{L} = \mathcal{L}_{M} + \alpha \cdot \mathcal{L}_{V} + \beta \cdot \mathcal{L}_{C}, 
\end{equation}
where $0\leq \alpha \leq1$ and $0\leq \beta \leq1$ are trade-off parameters.

In the inference process, we input the textual query into InterLLaVA. The system then generates outputs, including intent analysis, video prediction, or video moment prediction, along with explainability feedback.

\section{Experiments}\label{sec:exp}

\setlength{\tabcolsep}{2.5mm}  % Adjusts the horizontal padding between columns
\renewcommand\arraystretch{1.4}

\begin{table*}[t]    
    \centering
    \caption{Overall performance comparison of baselines. ``--'' indicates not applicable, while \textbf{bold} represents optimal performance.}   
    \newsavebox{\tabbb}
    \begin{lrbox}{\tabbb}
    \resizebox{\textwidth}{!}{
    \begin{tabular}{c|c||c||c|c||c|c}
    \toprule
    Types &  Methods& R@1 ↑  & 
    \begin{tabular}[c]{@{}c@{}}R@1 \\ IoU=0.5\end{tabular} ↑  & \begin{tabular}[c]{@{}c@{}}R@1 \\ IoU=0.7\end{tabular} ↑  & BLEU-4 ↑       & GPT-4 Score  ↑ \\ \hline
    \multirow{4}{*}{\begin{tabular}[c]{@{}c@{}}Video  Retrieval \end{tabular}}  & CLIP4Clip \citep{luo2021clip4clip}           &   42.0    &  --             &  --           & --            & --            \\ \cline{2-7} 
    & X-Pool \citep{gorti2022x}         &    47.3              &    --          &     --          & --            & --            \\ \cline{2-7} 
    & TS2-Net \citep{liu2022ts2}          &    45.0               &    --          &     --          & --            & --            \\ \cline{2-7} 
    & CLIP-ViP \citep{xue2023clip}                  & 50.4    &       --          &     --          & --            & --            \\ \hline \hline
    
    \multirow{4}{*}{\begin{tabular}[c]{@{}c@{}}Moment  Retrieval \end{tabular}}
    & XML \citep{lei2020tvr}        &  --       &     10.1       &  5.2        & --            & --            \\ \cline{2-7} 
    & TimeChat \citep{ren2024timechat}             &   --    &      15.4       &    8.1     & --            & --            \\ \cline{2-7} 
    & VTimeLLM \citep{huang2024vtimellm}             &   --    &      14.9       &    6.6     & --            & --            \\ \cline{2-7}   
    & GroundingGPT \citep{li2024groundinggpt}            &   --        &    10.7  &   6.3      & 0.0018           & 0.68         
    \\ \hline \hline
    
    \multirow{1}{*}{Interactive Video Corpus Retrieval}            
                                      & \textbf{InterLLaVA (Ours)}  & \textbf{54.6}  &\textbf{19.2} & \textbf{9.8} & \textbf{0.42}&\textbf{0.76}  \\
    \bottomrule
    \end{tabular}}
    \end{lrbox}
    \scalebox{0.9}{\usebox{\tabbb}}  
    \vspace{-0.15cm}
\label{tab:table_I}
\end{table*}

\begin{table}[t]
    \setlength{\tabcolsep}{7pt}
    \centering  
    \caption{Performance of different pre-retrieval modules.}  
    %\vspace{-0.3cm}  
    \begin{tabular}{c||c||c|c}  
    \toprule  
        Models  & R@1 ↑ & \begin{tabular}[c]{@{}c@{}}R@1 \\ IoU=0.5\end{tabular}  ↑ & \begin{tabular}[c]{@{}c@{}}R@1 \\ IoU=0.7 \end{tabular}    ↑    \\ \hline  
    CLIP4Clip \citep{luo2021clip4clip}   & 53.8   & 17.8 &  8.7   \\ \hline  
    
    TS2-Net \citep{liu2022ts2}  & 53.5 &  18.5 & 9.2      \\  \hline 
    X-Pool \citep{gorti2022x}  & 54.6 & 19.2 & 9.8  \\  
    \bottomrule
    \end{tabular}
    \vspace{-0.3cm}
    \end{table}
    
    \begin{table}[t]
    \setlength{\tabcolsep}{10pt}
    \centering  
    \caption{Multi-turn analysis of the proposed framework.}  
    %\vspace{-0.3cm}
    \label{tab:turns}  
    \begin{tabular}{c||c||c|c}  
    \toprule  
                    &  R@1  ↑         & \begin{tabular}[c]{@{}c@{}}R@1 \\ IoU=0.5\end{tabular} ↑  & \begin{tabular}[c]{@{}c@{}}R@1 \\ IoU=0.7\end{tabular}  ↑       \\ \hline  
    Turn 1\#             &   38.4     &   15.7         &     6.3                         \\ \hline  
    Turn 2\#                &  45.2        &    21.9       &     8.6                       \\ \hline  
    Turn 3\#                &   50.9        &   27.7       &    10.8                      \\ \hline  
    Turn 4\#                &  64.5       &    29.8       &   11.4                        \\ 
    \hline
    Turn 5\#&  65.4       &    31.5       &   15.8                        \\ 
    \hline
    Turn 6\#&  83.3       &    32.7       &   19.0                        \\ 
    \hline
    Turn 7\#&  87.5       &    34.9       &   19.7                        \\ 
    \bottomrule
    \end{tabular}
    %\vspace{-0.2cm}
\label{tab:table_II}  
\end{table}

\subsection{Experimental Settings}

\stitle{Datasets Splits.} 
Our datasets are split into three non-overlapping subsets, where 0.8, 0.1, and 0.1 are used for training, testing, and validation. Specifically, the training set comprises 10,016 videos and 89,329 textual queries, while the test set consists of 1,250 videos and 11,310 textual queries. The validation set comprises 1,250 videos and 10,979 textual queries. 

\stitle{Evaluation Metrics.} 
We employ two types of metrics to assess the proposed framework. For single-turn evaluation, we use R@1 to measure video retrieval performance, where ``1'' indicates that the relevant video is ranked first. R@1 IoU=\{0.5, 0.7\} is employed to assess video moment retrieval capability, with IoU=0.5 indicating that the IoU score between the localized moment and the ground truth must exceed 0.5. Metrics such as BLEU-4 and GPT-4 Score are deployed to evaluate text generation. We classify GPT-4 scores into four categories: highly relevant (1), moderately relevant (0.6), somewhat relevant (0.4), and irrelevant (0). Moreover, we conduct multi-turn performance based on the metrics as mentioned earlier, and any error between rounds will affect subsequent scores.

\stitle{Baselines.}  
For video retrieval, we selected the following four state-of-the-art models as baselines. We adopt their original setup, using both video and text as model inputs for the video retrieval task. CLIP4Clip \citep{luo2021clip4clip} uses CLIP to extract the frame features and the text features, and then uses the mean pooling to aggregate the feature of all frames for video representation. X-Pool \citep{gorti2022x} adopts text-conditioned pooling to aggregate visual features. TS2-Net \citep{liu2022ts2} proposes different token shift operations in ViT to learn short-term temporal dependencies across locally adjacent frames. CLIP-ViP \citep{xue2023clip} proposes a CLIP-based video post-pretraining framework, enabling efficient transfer of an image–text pretrained model to video tasks. For video moment retrieval, we selected four methods as baselines. XML \citep{lei2020tvr} designs a ConvSE module that detects start and end edges in 1D similarity signals using two convolution filters. TimeChat \cite{ren2024timechat} proposes a time-sensitive multimodal large language model for long video understanding and precise temporal localization. VTimeLLM \citep{huang2024vtimellm} uses a boundary-aware three-stage strategy to align image–text features, enhance temporal boundaries via multi-event video–text QA, and refine temporal reasoning through dialogue-based instruction tuning. GroundingGPT \citep{li2024groundinggpt} proposes a time-sensitive multimodal large language model for long video understanding and precise temporal localization. 

\stitle{Implementation Details.}
We employ a pre-trained ViT-G/14 from EVA-CLIP \citep{sun2023eva} and the sliding video Q-Former \citep{ren2024timechat} as the image encoder, with LLaMA-2 (7B) \citep{touvron2023llama} as the language model backbone. We train InterLLaVA using the AdamW optimizer with an initial learning rate of 3e-5 and weight decay of 1e-6 in training phases 1 and 2. Fine-tuning was performed on IVCR-200K for five epochs with a batch size of 32. As depicted in Figure \ref{fig:framework}, the parameters of ViT and LLM remained frozen, while those of the image Q-Former, video Q-Former, and linear layer were tuned. For video retrieval, 64 frames are used, while for moment retrieval, 96 frames are used. In addition, the trade-off parameter $\alpha$ an $\beta$ in Eq.~(\ref{eqn:loss}) are set to $0.01$.

\subsection{Overall Performance Comparison}

To assess the challenges posed by the IVCR-200K dataset, we conducted a comprehensive evaluation of multiple models across its tasks, including our benchmark. Table \ref{tab:table_I} presents a comparison of InterLLaVA against state-of-the-art video retrieval and moment retrieval methods.

\stitle{Overall Observations.} 
1) Notice that the IVCR task presents significant challenges in the field of video retrieval. While existing traditional models have achieved notable success in single tasks such as video retrieval and video moment retrieval, they fall short compared to InterLLaVA in terms of considering the importance of flexibly adjusting retrieval strategies based on retrieval intent. This limitation restricts the flexibility and adaptability of video retrieval to some extent. 
2) For video moment retrieval, compared to traditional methods (e.g., XML), multimodal large language models (e.g., GroundingGPT) have achieved superior performance in moment localization. Their advantage lies in the ability to capture richer contextual information. 
3) Moreover, the CLIP-based models, CLIP-ViP and X-Pool, have demonstrated excellent performance on the video retrieval task. This proves their ability to align key textual and video information more effectively. 

\subsection{Robustness Analysis}

In this section, we will delve into the proposed framework from two perspectives: the retrieval module and the multi-turn analysis. We will examine the functionality of the retrieval module within the framework and evaluate the performance of multi-turn dialogue.

\stitle{Retrieval Module.} 
We validate the effectiveness of interactive retrieval modeling by substituting different video retrieval models in Table \ref{tab:table_II}, which demonstrates the following points:
1) Upon comparing Tables~\ref{tab:table_I}, \ref{tab:table_II}, it becomes apparent that, for the video retrieval task, X-Pool demonstrates significantly greater performance improvements compared to the TS2-Net. 2) In contrast, for the video moment retrieval task, CLIP4Clip exhibits slightly diminished performance, suggesting that the underlying video retrieval model influences InterLLaVA's video localization capabilities. Overall, these observations empirically validate the effectiveness of video retrieval models and large language models in modeling interactive retrieval.

\stitle{Multi-Turn Analysis.} 
To evaluate the model's effectiveness, we compared its performance across different dialogue turns. As shown in Table \ref{tab:turns}, retrieval performance for both video and moment search improves significantly with additional turns, highlighting the pivotal role of contextual learning in enhancing retrieval and localization across multiple interactions.

\setlength{\tabcolsep}{6pt}
\begin{table}[t]
    \centering
    \caption{Comparing InterLLaVA performance across different data distributions.}
    \label{tab:data}
    \begin{tabular}{c|c|c|c|c}
    \toprule
    Training  & Category & R@1 ↑ & \begin{tabular}[c]{@{}c@{}}R@1  \\ IoU=0.5\end{tabular}  ↑ & \begin{tabular}[c]{@{}c@{}}R@1 \\ IoU=0.7 \end{tabular} ↑       \\ \hline  
    
    2K  & Movies & 21.36 & 6.21 & 2.24  \\ \hline
    20K & TV shows & 26.73 & 8.24 & 3.98 \\ \hline
    1K  & ALL& 36.6 & 8.49 & 4.94 \\ \hline
    6K  & ALL & 45.7 & 11.31 & 5.8 \\ \hline
    
    20K & ALL & 47.96 & 11.52 & 6.45 \\ \hline
    60K & ALL & 51.86 & 14.28  & 7.13  \\
    \bottomrule
    \end{tabular}
    \vspace{-0.43cm}
\end{table}

\subsection{Training Data Distribution Analysis} 

InterLLaVA is trained on four different data modes, with the number of training samples ranging from 1K to 60K. Table \ref{tab:data} summarizes the performance evaluation results for all training samples. We observed the following points: 1) In low-sample scenarios, particularly when the sample size is less than 6K, InterLLaVA's accuracy is significantly limited (36.6 vs 54.86 for R@1), showing much lower performance compared to conditions with larger sample sizes. 2) We further explored training the model using samples from a single category, such as movies. The experimental results indicate that compared to training on data of the same scale but with more diverse categories, InterLLaVA's video and moment retrieval performance decreased by 21.23\% and 3.28\% in R@1 and R@1 IoU=0.5, respectively. This result aligns with expectations, as training on more diverse categories allows the model to capture richer features and enhance its generalization ability. 

\section{Details of Investigation of User Search Behavior Feedback}\label{sec:inves} 

To verify users' search needs for interactive retrieval, we design a user search behavior feedback survey. The survey covers several aspects: user demographics (such as age, gender, occupation), video search habits, video search usage (whether users prefer to search for video moment directly), evaluation of search effectiveness (including the accuracy and usability of current video search systems, user satisfaction, and expectations for conversational search tools), and issues and improvement suggestions (open-ended questions for users to share problems and suggestions when using interactive search). To better accommodate the habits and preferences of users from different countries, we implemented a diversified survey approach. On the Amazon Mechanical Turk\footnote{https://www.mturk.com/} platform, we conducted an online survey specifically designed to target an international audience, successfully engaging 500 participants from diverse regions. In China, we utilized the Wenjuanxing\footnote{https://www.wjx.cn/} platform to reach and collect feedback from 500 Chinese users accurately. Additionally, we organized offline paper surveys. Over the weekends, we gathered insights from an additional 500 participants in high-traffic areas, including university campuses, shopping malls, parks, popular tourist destinations, and subway stations.

% \section{Limitations}~\label{sec:lim} 

% The IVCR-200K dataset is constrained by the depth of manual annotation and the diversity of real-world data types. It needs to be expanded to cover a wider array of interactive retrieval scenarios, including complex analogy searches, diverse contextual searches, and fine-grained interactive search requirements. Additionally, the current model does not achieve seamless integration of video retrieval and moment retrieval into a unified, efficient end-to-end system. There is considerable potential for improvement in areas such as temporal video modeling, accurately capturing user retrieval intent, and the natural and fluid execution of multi-round dialogues.

\section{Conclusions}\label{sec:conclus}

In this paper, we propose a more realistic task to establish an ``interaction" between the retrieval system and the user, which involves multi-turn, conversational, and realistic interactions. To facilitate research on this challenging task, we introduce a dataset and framework designed to serve this novel purpose. Notably, the IVCR-200K dataset is a high-quality, bilingual, multi-turn, conversational, and abstract semantic dataset that supports both video and moment retrieval. Our framework is based on MLLMs, which provide more explainable solutions to help users' interaction modes. The extensive experiments demonstrate the effectiveness of IVCR-200K and the proposed framework.
Moving forward, we plan to expand the scope of this research by increasing the dataset size and model parameters. Additionally, we will endeavor to develop more sophisticated models to enhance the model's capabilities, taking into account the challenges posed by interactive retrieval.

\section*{Acknowledgments} 

This research is supported by the National Natural Science Foundation of China (No. 62272156), the Key Technology Research and Development Program of Hunan Province (No. 2024QK2010), and the Natural Science Foundation of Hunan Province (No. 2024JJ6435).

\clearpage
\newpage

\bibliographystyle{ACM-Reference-Format}
\bibliography{sample-base}

\end{document}